\documentclass[sigconf]{acmart}
\usepackage{multirow}
\usepackage{multicol}
\usepackage[most]{tcolorbox}
\tcbset{
  mypromptbox/.style={
    colback=gray!10,       % 背景色
    colframe=gray!150,     % 边框色
    arc=1mm,               % 圆角
    boxrule=1.2pt,         % 边框粗细
    left=3pt, 
    right=3pt, 
    top=3pt, 
    bottom=3pt,
    fonttitle=\bfseries,
    coltitle=black,
  }
}
\usepackage{pifont}  % 放在导言区
\newcommand{\cmark}{\textcolor{green!50!black}{\ding{51}}}  % Deep Green Checkmark
\newcommand{\xmark}{\textcolor{red}{\ding{55}}}  % Red Cross

\AtBeginDocument{%
  }

\copyrightyear{2025}
\acmYear{2025}
\acmDOI{XXXXXXX.XXXXXXX}
\acmConference[ACM]{Proceedings of the 33th ACM International Conference on Multimedia}{2025}{Manuscript}
\acmBooktitle{Proceedings of the 30th ACM International Conference on Multimedia (MM '22), October 10--14, 2022, Portugal, Lisbon}
\acmISBN{978-1-4503-XXXX-X/18/06}

\acmSubmissionID{xxxx}

\begin{document}
\title{Automatic Synthesis of High-Quality Triplet Data for Composed Image Retrieval}

\author{Haiwen Li}
\affiliation{%
  \institution{Beijing University of Posts and Telecommunications}
  \city{Beijing}
  \country{China}}
\email{lihaiwen@bupt.edu.cn}

\author{Delong Liu}
\affiliation{%
  \institution{Beijing University of Posts and Telecommunications}
  \city{Beijing}
  \country{China}}
\email{liudelong@bupt.edu.cn}

\author{Zhaohui Hou}
\affiliation{%
  \institution{SenseTime}
  \city{Beijing}
  \country{China}}
\email{houzhaohui@sensetime.com}

\author{Zhicheng Zhao}
\affiliation{%
  \institution{Beijing University of Posts and Telecommunications}
  \city{Beijing}
  \country{China}}
\email{zhaozc@bupt.edu.cn}

\author{Fei Su}
\affiliation{%
  \institution{Beijing University of Posts and Telecommunications}
  \city{Beijing}
  \country{China}}
\email{sufei@bupt.edu.cn}

\begin{abstract}
As a challenging vision-language (VL) task, Composed Image Retrieval (CIR) aims to retrieve target images using multimodal (image+text) queries. Although many existing CIR methods have attained promising performance, their reliance on costly, manually labeled triplets hinders scalability and zero-shot capability. To address this issue, we propose a scalable pipeline for automatic triplet generation, along with a fully synthetic dataset named Composed Image Retrieval on High-quality Synthetic Triplets (CIRHS). Our pipeline leverages a large language model (LLM) to generate diverse prompts, controlling a text-to-image generative model to produce image pairs with identical elements in each pair, which are then filtered and reorganized to form the CIRHS dataset. In addition, we introduce Hybrid Contextual Alignment (CoAlign), a novel CIR framework, which can accomplish global alignment and local reasoning within a broader context, enabling the model to learn more robust and informative representations. By utilizing the synthetic CIRHS dataset, CoAlign achieves outstanding zero-shot performance on three commonly used benchmarks, demonstrating for the first time the feasibility of training CIR models on a fully synthetic dataset. Furthermore, under supervised training, our method outperforms all the state-of-the-art supervised CIR approaches, validating the effectiveness of our proposed retrieval framework. The code and the CIRHS dataset will be released soon.
\end{abstract}

%    <concept>
%        <concept_id>10002951.10003317.10003371.10003386.10003387</concept_id>
%        <concept_desc>Information systems~Image search</concept_desc>
%        <concept_significance>500</concept_significance>
%        </concept>
%  </ccs2012>
% \ccsdesc[500]{Information systems~Image search}


\begin{teaserfigure}
\centering
\includegraphics[width=0.96\textwidth]{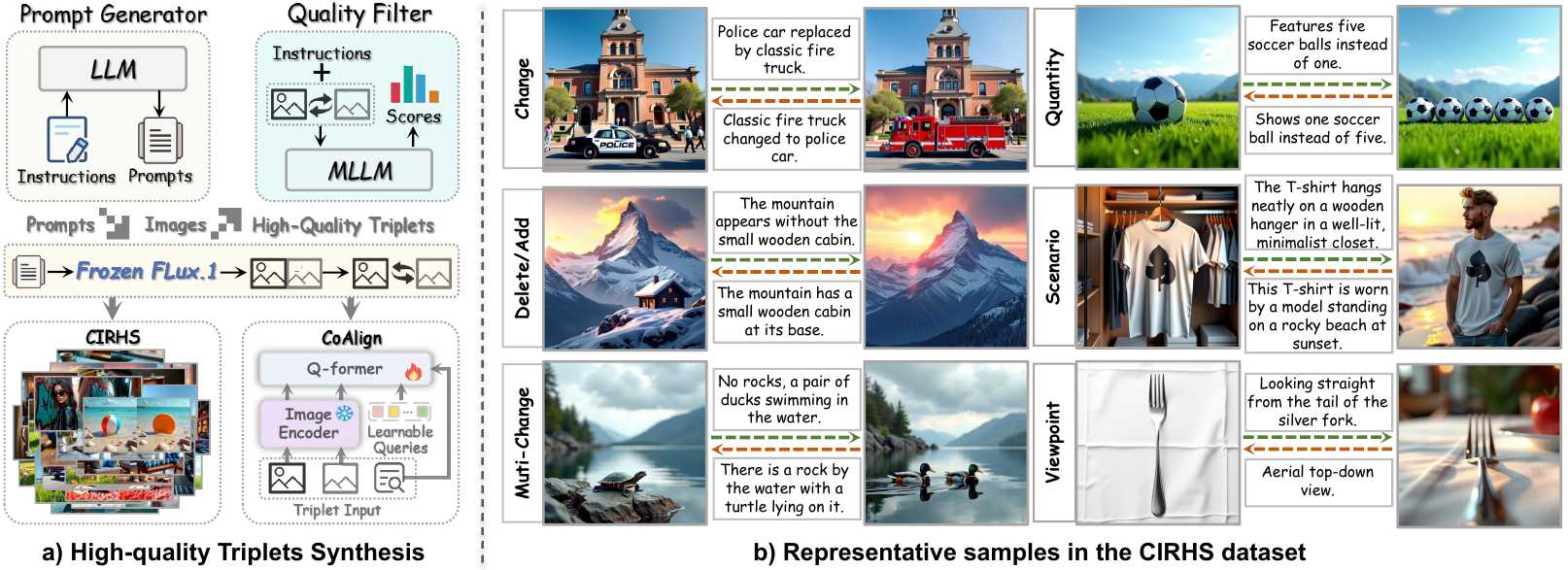}
\caption{\textbf{Overview of our key innovations.} (a) Illustration of the proposed automatic triplet synthesis pipeline, along with our proposed training framework CoAlign. (b) Representative samples from our CIRHS dataset, covering various real-world scenes and objects, as well as diverse editing operations including object or scene change, quantity variation, and viewpoint shift, etc.}
% \caption{Overview of our key innovations. (a) Illustration of the proposed automatic triplet construction pipeline, along with the high-quality synthetic dataset CIRHS and our proposed training framework CoAlign. (b) Representative samples from our CIRHS dataset, covering a wide range of real-world scenes and objects, as well as diverse editing operations including object substitution, scene change, quantity variation, viewpoint shift, etc.}
\label{fig:intro}
\end{teaserfigure}


\maketitle

\section{Introduction}
Composed Image Retrieval (CIR)~\cite{vo2019composing,fashioniq,cirr} has attracted increasing attention in recent years, aiming to retrieve target images based on a query consisting of a reference image and a relative caption. By integrating information from both modalities, CIR can attain more precise and flexible searches, and provide a superior user experience compared to unimodal retrieval~\cite{unimodal}. With the emergence of large-scale vision-language pretraining (VLP) models~\cite{clip,alayrac2022flamingo,blip2}, CIR has made significant progress and found numerous applications, especially in e-commerce and web search. 

Existing supervised CIR approaches~\cite{artemis,clip4cir,blip4cir,cala,sprc} heavily rely on manually annotated triplets, which are both time-consuming and labor-intensive to construct. Moreover, due to the limited scale of available datasets such as FashionIQ~\cite{fashioniq} with only 46.6k triplets and CIRR~\cite{cirr} with just 28.8k, these methods suffer from poor generalization performance. As a result, several studies~\cite{pic2word,serealcirco} present Zero-Shot Composed Image Retrieval (ZS-CIR). Early approaches~\cite{pic2word,serealcirco} employ inversion networks~\cite{textinv,personalize} trained on massive image-text pairs. However, these methods have inherent task discrepancy~\cite{rtd}, making them suboptimal solutions. Additionally, some training-free approaches~\cite{cirevl,ldre} introduce large language model (LLM) reasoning into ZS-CIR. While promising, they often fail to capture fine-grained visual details and the high complexity of model architecture makes it infeasible to conduct domain-specific fine-tuning, thus limiting their applicability in customized scenarios. Recent studies~\cite{covr,compodiff,vista} have designed automated pipelines for creating large-scale triplet datasets, and they also unify the model architecture for both ZS-CIR and supervised CIR. Compared with previous methods, this line of work enables more robust generalization while preserving the ability to fine-tune on domain-specific datasets. However, they exhibit two obvious limitations: (1) The relative captions cover only a narrow range, primarily focusing on object or scene substitution, which results in a lack of diversity. (2) The images produced by image editing~\cite{pix2pix,ldm} are often of low quality, with severe artifacts and poor semantic consistency.

To deal with the above issues, we propose a scalable pipeline for automatic triplet synthesis as illustrated in Figure~\ref{fig:intro}(a), which generates high-quality CIR triplets in three stages. In Stage 1, an LLM is employed to generate numerous textual quadruplets, each consisting of two image captions as well as two relative captions that describe how one image can be transformed into the other. Guided by a carefully crafted instruction with randomized elements, the LLM-generated image captions cover a wide range of real-world objects and scenes, while the relative captions include diverse editing operations such as object replacement, attribute modification, and scene composition, thus effectively relieving the first limitation mentioned before. However, textual quadruples alone are insufficient for training CIR models, as the LLM-generated image captions need to be converted into corresponding images. 

Therefore, in Stage 2, we focus on how to obtain high-quality images corresponding to the textual quadruples. A straightforward solution is to use a text-to-image generative model (T2I-GM) to synthesize image independently from each caption. However, this will inevitably faces a major drawback: CIR requires consistency in shared elements between the reference and target images, while independent generation often results in uncontrollable visual discrepancies. Fortunately, we notice that T2I-GMs are capable of preserving strong intra-image consistency. Consequently, we first combine the two image captions in each textual quadruple into a single prompt using a predefined template. This prompt is then fed into the T2I-GM to produce an image containing two semantically related sub-images, which are cropped respectively to serve as the reference image and the target image. 

Stage 3 performs data filtering, where the generated images and relative captions are reorganized into triplets. Each triplet is scored by a multimodal large language model (MLLM) based on three criteria: image quality, image–caption fidelity and CIR task alignment. Afterwards, we compute a weighted sum of the three scores and discard the bottom 15\% of the triplets, resulting in a high-quality and fully synthetic dataset of 534k triplets, namely Composed Image Retrieval on High-Quality Synthetic Triplets (CIRHS), with representative samples shown in Figure~\ref{fig:intro}(b). Moreover, we propose a unified framework named Hybrid Contextual Alignment (CoAlign), which is applicable to both supervised and zero-shot CIR. CoAlign optimizes the model within a broader context, combining global and local objectives to learn more robust and fine-grained representations. Extensive experiments on three popular benchmarks validate the effectiveness of our method under both CIR and ZS-CIR settings, as well as the feasibility of training CIR models using purely synthetic data. To sum up, our contributions are threefold:
\begin{itemize}
\item We propose a scalable pipeline for automatic CIR triplet synthesis, tackling previous limitations such as low image generation quality, poor semantic consistency, and the lack of diversity in relative captions. With this pipeline, we obtain a large-scale, fully synthetic CIR dataset named CIRHS, which consists of 534K high-quality triplets.
\item We propose a novel CIR framework, Hybrid Contextual Alignment (CoAlign), which optimizes the model within a broader context by combining global alignment and local reasoning. It is simple yet effective and can enhance the robustness of learned representations.
\item Experiments on three benchmarks show the superior performance under both CIR and ZS-CIR settings. To the best of our knowledge, this is also the first work to verify the feasibility of training CIR models solely on synthetic data.
\end{itemize}

\section{Related Work}
\subsection{Composed Image Retrieval}
CIR is primarily evaluated on the fashion domain~\cite{fashioniq} and real-world scenarios~\cite{cirr,serealcirco}. Mainstream approaches~\cite{artemis,clip4cir,blip4cir,cala,sprc}  leverage the strong cross-modal alignment capabilities of VLP models, and apply early or late fusion to integrate the two modalities in the composed query. Recently, zero-shot CIR has been widely explored, with textual inversion~\cite{pic2word,serealcirco,textinv,personalize} emerging as a key technique. It maps an image to a pseudo-word token, which is combined with a relative caption and encoded via a text encoder. LinCIR~\cite{lincir} further improve this approach by training the inversion network purely on text, boosting both efficiency and performance. Another line of work~\cite{cirevl,ldre} harnesses the strong reasoning capacity of LLMs to generate target captions and reformulates CIR as a text-to-image retrieval task. While effective, its complex model architecture hinders domain-specific fine-tuning and deployment in resource-constrained environments. Owing to the scarcity of manually annotated triplet datasets, recent efforts attempt to construct synthetic triplets automatically. Some approaches~\cite{covr,case} collect similar image pairs from public databases and generate relative captions by hand-crafted rules or LLMs. CompoDiff~\cite{compodiff} and VISTA~\cite{vista} synthesize target images via image editing~\cite{pix2pix,ldm}, resulting in large-scale synthetic datasets, but their performance is highly constrained by the quality of the generated data. Distinctively, the triplets generated by our method include relative captions covering diverse editing operations, together with high-quality, photorealistic reference and target images. This enables CIR models to learn more comprehensive representations.

\begin{figure*}
  \includegraphics[width=0.92\textwidth]{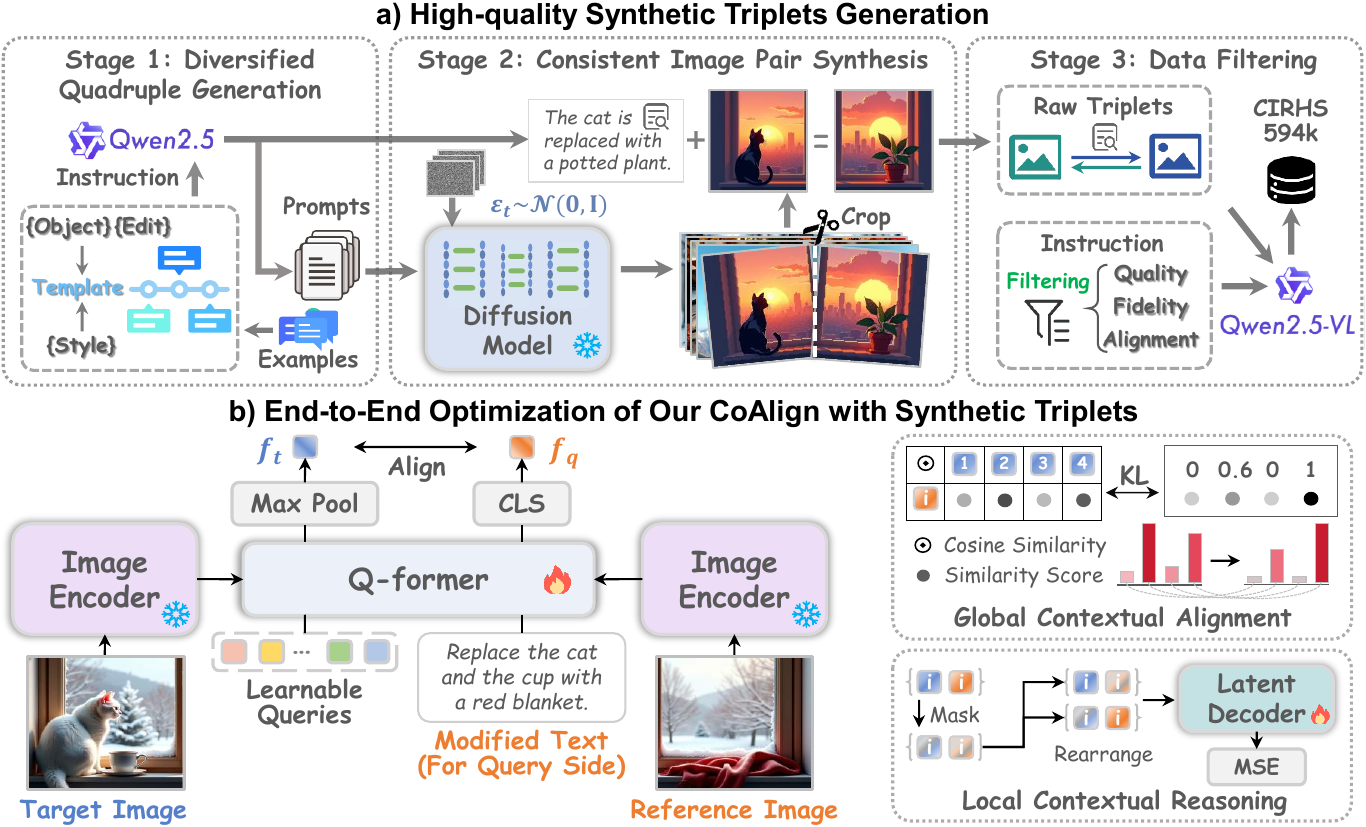}
  \caption{\textbf{Overall framework of our method.} (a) The triplet synthesis pipeline consists of three stages. In the first stage, an LLM is guided to generate diverse textual quadruples. Subsequently, consistent image pairs are synthesized using the quadruples and reorganized into triplet form. The final stage leverages an MLLM to filter out low-quality samples. (b) \textbf{The model architecture of CoAlign.} The left side illustrates the encoding process of the query and target using different encoding modes, while the right shows the global and local optimization objectives employed by CoAlign.}
  \label{fig:main}
\end{figure*}

\subsection{Text-to-Image Generative Models}
The development of image generation has undergone transformative shifts, evolving from early GAN-based methods~\cite{gan,wgan} to more complex multimodal frameworks. Currently, diffusion models~\cite{diff1,diff2} have emerged as the prevailing paradigm, showing exceptional performance in tasks such as text-to-image synthesis~\cite{text1,text2}, image-to-image translation~\cite{image1,image2,image3}, and controllable generation~\cite{con1,con2,con3}. In the domain of text-to-image synthesis, the introduction of latent diffusion models (LDM)~\cite{ldm} significantly enhances text-image fidelity while reducing computational costs through latent space operations. This also give rise to excellent works supporting high-resolution image generation, such as Stable Diffusion~\cite{sdxl} and DALL-E 2~\cite{dalle2}. Recently, the integration of transformer architectures~\cite{attention} in Diffusion Transformers (DiT) \cite{dit} has further improved scalability, leading to the development of advanced models like PixArt~\cite{pix1} and Flux~\cite{flux}. These models adopt flow-matching~\cite{fm} to achieve state-of-the-art generation quality, facilitating the synthesis of high-quality images with highly aligned text and multi-object or multi-subgraph structures. Benefiting from this, we transform the challenge of "generating images similar to the reference images" into a more tractable problem "generating images that contain identical elements in two subgraphs", thereby enabling scalable image pair synthesis when constructing CIR triplets.
% 最近，Diffusion Transformers (DiT)进一步提高了可扩展性，使得
\section{Methodology}
We introduce our method in two subsections: \ref{synthesis} presents our automatic triplet synthesis pipeline, including textual quadruple generation, consistent image pair synthesis, and data filtering, while \ref{coalign} describes our proposed CIR framework, CoAlign, detailing its model architecture, optimization strategy, and inference workflow.
\subsection{High-Quality Triplet Synthesis for CIR}
\label{synthesis}
% 首先，我们分三个stage来合成triplet dataset，具体下文所述
\subsubsection{Diverse Quadruple Generation}
Considering that end-to-end generation of CIR triplets is highly challenging. Therefore, in Stage 1, we generate the corresponding textual counterparts instead. To be specific, an instruction template $\mathcal P(object,edit,style)$ is designed to guide an LLM~\cite{qwen2} $g_{llm}(\cdot)$ in generating textual quadruples as shown in Equation~\ref{eq:llm}. The three parameters of $\mathcal P$ are respectively sampled from predefined sets covering a wide range of real-world objects, editing operations, and image styles.
\begin{equation}
g_{llm}(p)\rightarrow\langle {C_{I_r}}, C_{r\rightarrow t}, C_{t\rightarrow r}, C_{I_t}\rangle.
\label{eq:llm}
\end{equation}
where the reference caption $C_{I_r}$ and the target caption $C_{I_t}$ are used to synthesize the corresponding images $I_r$ and $I_t$, sharing at least one semantic entity, while the relative caption $C_{r\rightarrow t}$ captures the modification from $I_r$ to $I_t$. Additionally, the inverse caption $C_{t\rightarrow r}$ is introduced to describe the change from $I_t$ to $I_r$, enabling bidirectional triplet construction and thereby improving efficiency. The instruction $p\sim \mathcal{P}$ varies with each input, encompassing various common objects and editing operations such as object composition and scene changes. Furthermore, style information is embedded in $C_{I_r}$ and $C_{I_t}$ for image generation across multiple domains. An illustration of the instruction template $\mathcal{P}$ is shown below.

\begin{tcolorbox}[mypromptbox]
Using the elements: \{\textcolor{blue}{suggested objects}\}. \{\textcolor{blue}{editing operations}\}. \{\textcolor{blue}{image styles}\}, please help me generate a quadruple that meets the requirements of CIR.
\end{tcolorbox}

\subsubsection{Consistent Image Pair Synthesis} 
After obtaining the textual quadruplets, a T2I-GM can synthesize $I_r$ and $I_t$ using $C_{I_r}$ and $C_{I_t}$, respectively. However, the stochastic nature of generative models may undermine the consistency of shared elements across the two separately generated images. That is, $I_r$ and $I_t$ may differ significantly, making them unsuitable for constructing CIR triplets. 

In contrast, we leverage the inherent consistency of generative models, i.e., \textit{they have the ability to generate identical elements within a single image.} To this end, we define a prefix to specify the desired image layout and integrate $C_{I_r}$ and $C_{I_t}$ into a single prompt as shown below, which is then fed into the T2I-GM~\cite{flux} to produce a single image containing two side-by-side sub-images. The reference image $I_r$ and the target image $I_t$ are finally obtained by cropping the left and right parts, respectively.

\begin{tcolorbox}[mypromptbox, halign=center]
HD 4k, square grid layout... Left: \{\textcolor{blue}{$C_{I_r}$}\}, Right: \{\textcolor{blue}{$C_{I_t}$}\}.
\end{tcolorbox}

To maximize the utility of textual quadruples, we synthesize $n$ image pairs for every $(C_{I_r}, C_{I_t})$, yielding $\{(I_r^i, I_t^i)\}_{i=1}^n$. By combining these images with the relative captions, we obtain $2n$ CIR triplets, i.e., $\{(I_r^i,C_{r\rightarrow t},I_t^i)\}_{i=1}^n\cup\{(I_t^i,C_{t\rightarrow r},I_r^i)\}_{i=1}^n$. Additionally, we introduce an identifier, namely triplet identity (TID), and assign the same TID to all triplets sharing the same relative caption. Triplets with the same TID exhibit a certain degree of similarity, making label smoothing~\cite{soft} possible during training.

\subsubsection{Data Filtering} 
To refine the raw synthesized triplets, in Stage 3, we employ a multimodal large language model (MLLM)~\cite{qwen2-vl} to score them across three aspects on a scale from $1$ to $10$: (1) image quality of both $I_r$ and $I_t$, including the clarity, noise, artifacts, etc. (2) image–caption fidelity (e.g. $I_r\leftrightarrow C_{I_r}$), and (3) the alignment of CIR task ($I_{r}+C\rightarrow I_t$). The average score is computed via a weighted sum and a threshold $\alpha$ is then applied to filter out low-quality triplets, which accounts for approximately 15\%.

Using this pipeline, we build the large-scale CIRHS dataset, which contains 534k high-quality synthetic triplets. Experiments will verify the scalability and effectiveness of this pipeline. Additional details on data synthesis, including the full prompts used at each stage and more visualizations, can be found in Appendix A.
% In summary, our method is both scalable and efficient，可以很容易应用到大规模triplet dataset的自动生成。基于此我们得到了CIRHS数据集，包含了534k个高质量三元组。关于数据合成更多的细节，比如每个stage采取的完整的prompt以及数据筛选的细节，可以在附录中找到

\subsection{End-to-End Optimization for Composed Image Retrieval Using Synthetic Triplets}
\label{coalign}
We introduce our CoAlign model architecture first and then describe how it is trained on triplet data.
\subsubsection{CoAlign Model Architecture} 
As illustrated in Figure~\ref{fig:main}(b), inspired by BLIP-2~\cite{blip2}, our model consists of a frozen image encoder and a lightweight Querying Transformer (Q-Former), which incorporates learnable queries for efficient multimodal feature extraction. CoAlign reuses the Q-Former's two distinct encoding modes: image-grounded encoding (jointly conditioned on visual and textual inputs) and pure image encoding (visual-only processing). 

Given an input triplet $\langle I_{r}, C, I_{t} \rangle$, the query side uses the frozen image encoder to extract features from the reference image $I_r$. The resulting visual features, together with the modified text $C$, are forwarded to the Q-Former. The output [CLS] token of the modified text is then passed through a textual projection layer to produce the query feature $f_q\in\mathbb R^d$. Similarly, on the target side, the frozen image encoder processes the target image $ I_{t}$, generating visual features that are then passed through the Q-Former in its pure image encoding mode. The output token embeddings corresponding to the learnable queries of the Q-Former undergo max-pooling across the sequence dimension, followed by a visual projection layer, to obtain the target feature $f_t\in\mathbb R^d$.

\subsubsection{Hybrid Contextual Alignment}
To achieve comprehensive alignment between the composed query and its corresponding target images, CoAlign jointly conducts optimization from both global and local perspectives.

\begin{table*}[t]
\small
\centering
\caption{\textbf{Performance comparison with existing supervised CIR methods.} The best results are marked in bold, and the second-best results are underlined. $\ddagger$ indicates that the method is pretrained on its own constructed triplet dataset.}
\begin{tabular}{c c c c c | c c c | c c | c c | c c | c c }
\toprule
& & \multicolumn{6}{c}{\textbf{CIRR}} & \multicolumn{8}{c}{\textbf{FashionIQ}} \\
\multirow{2}{*}{\textbf{Method}} & \multirow{2}{*}{\textbf{Ref.}} & \multicolumn{3}{c}{Recall@K} & \multicolumn{2}{c}{Recall$_{s}$@K} & \multirow{3}{*}{Avg.} & \multicolumn{2}{c}{Dress} & \multicolumn{2}{c}{Shirt} & \multicolumn{2}{c}{Toptee} & \multicolumn{2}{c}{Average} \\ 
\cmidrule(lr){3-5} \cmidrule(lr){6-7} \cmidrule(lr){9-10} \cmidrule(lr){11-12} \cmidrule(lr){13-14} \cmidrule(lr){15-16}
& & K=1 & K=5 & K=10 & K=1 & K=3 & & R@10 & R@50 & R@10 & R@50 & R@10 & R@50 & R@10 & R@50 \\
\midrule
\multicolumn{1}{l}{TIRG~\cite{vo2019composing}} & \multicolumn{1}{l}{CVPR19} & 14.61 & 48.37 & 64.08 & 22.67 & 65.14 & 35.52 & 14.87 & 34.66 & 18.26 & 37.89 & 19.08 & 39.62 & 17.40 & 37.39 \\
\multicolumn{1}{l}{MAAF~\cite{maaf}} & \multicolumn{1}{l}{Arxiv20} & 10.31 & 33.03 & 48.30 & 21.05 & 61.60 & 27.04 & 23.80 & 48.60 & 21.30 & 44.20 & 27.90 & 53.60 & 24.30 & 48.80 \\
\multicolumn{1}{l}{CIRPLANT~\cite{cirr}} & \multicolumn{1}{l}{ICCV21} & 19.55 & 52.55 & 68.39 & 39.20 & 79.49 & 45.88 & 17.45 & 40.41 & 17.53 & 38.81 & 61.64 & 45.38 & 18.87 & 41.53 \\
\multicolumn{1}{l}{ARTEMIS~\cite{artemis}} & \multicolumn{1}{l}{ICLR22} & 16.96 & 46.10 & 61.31 & 39.99 & 75.67 & 43.05 & 27.16 & 52.40 & 21.78 & 43.64 & 29.20 & 53.83 & 26.05 & 50.29 \\
\multicolumn{1}{l}{CLIP4CIR~\cite{clip4cir}} & \multicolumn{1}{l}{CVPR22} & 38.53 & 69.98 & 81.86 & 68.19 & 94.17 & 69.09 & 33.81 & 59.40 & 39.99 & 60.45 & 41.41 & 65.37 & 38.32 & 61.74 \\
\multicolumn{1}{l}{TG-CIR~\cite{tgcir}} & \multicolumn{1}{l}{MM23} & 45.25 & 78.29 & 87.16 & 72.84 & 95.13 & 75.57 & 45.22 & 69.66 & 52.60 & 72.52 & 56.14 & 77.10 & 51.32 & 73.09\\
\multicolumn{1}{l}{Re-ranking~\cite{candidate}} & \multicolumn{1}{l}{Arxiv23} & 50.55 & 81.75 & \underline{89.78} & 80.04 & 96.58 & 80.90 & 48.14 & 71.43 & 50.15 & 71.25 & 55.23 & 76.80 & 51.17 & 73.13 \\
\multicolumn{1}{l}{BLIP4CIR+Bi~\cite{blip4cir}} & \multicolumn{1}{l}{WACV24} & 40.15 & 73.08 & 83.88 & 72.10 & 95.93 & 72.59 & 42.09 & 67.33 & 41.76 & 64.28 & 46.61 & 70.32 & 43.49 & 67.31\\
\multicolumn{1}{l}{CASE$^\ddagger$~\cite{case}} & \multicolumn{1}{l}{AAAI24} & 48.68 & 79.98 & 88.51 & 76.39 & 95.86 & 78.19 & 47.44 & 69.36 & 48.48 & 70.23 & 50.18 & 72.24 & 48.79 & 70.68 \\
\multicolumn{1}{l}{CoVR-BLIP$^\ddagger$~\cite{covr}} & \multicolumn{1}{l}{AAAI24} & 49.69 & 78.60 & 86.77 & 75.01 & 93.16 & 76.81 & 44.55 & 69.03 & 48.43 & 67.42 & 52.60 & 74.31 & 48.53 & 70.25 \\
\multicolumn{1}{l}{CompoDiff$^\ddagger$~\cite{compodiff}} & \multicolumn{1}{l}{TMLR24} & 32.39 & 57.61 & 77.25 & 67.88 & 94.07 & 62.75 & 38.39 & 51.03 & 41.68 & 56.02 & 45.70 & 57.32 & 39.81 & 51.90 \\
\multicolumn{1}{l}{CaLa~\cite{cala}} & \multicolumn{1}{l}{SIGIR24} & 49.11 & 81.21 & 89.59 & 76.27 & 96.46 & 78.74 & 42.38 & 66.08 & 46.76 & 68.16 & 50.93 & 73.42 & 46.69 & 69.22 \\
\multicolumn{1}{l}{SPRC~\cite{sprc}} & \multicolumn{1}{l}{ICLR24} & \underline{51.96} & \underline{82.12} & 89.74 & \underline{80.65} & \underline{96.60} & \underline{81.39} & \underline{49.18} & \textbf{72.43} & \underline{55.64} & \underline{73.89} & \textbf{59.35} & \underline{78.58} & \underline{54.72} & \underline{74.97} \\
\midrule
\multicolumn{1}{l}{\textbf{CoAlign (Ours)}} & - & \textbf{54.07} & \textbf{83.81} & \textbf{91.13} & \textbf{80.87} & \textbf{97.04} & \textbf{82.34} & \textbf{49.43} & \underline{72.04} & \textbf{56.48} & \textbf{75.61} & \underline{58.85} & \textbf{78.99} & \textbf{54.92} & \textbf{75.55} \\
\bottomrule
\end{tabular}
\label{supervised-sota}
\end{table*}

\textbf{Global Contextual Alignment:} In conventional contrastive learning~\cite{moco,infonce}, optimization is based on the diagonal elements of the predicted similarity matrix. However, this paradigm becomes suboptimal for the proposed CIRHS dataset, where each query 
may correspond to multiple target images (same TID). To address this, CoAlign integrates distribution matching~\cite{cmpm,irra} and label smoothing~\cite{soft} to perform global contextual alignment, enabling the model to extract useful information within a broader context, i.e. the entire similarity matrix, rather than relying solely on the diagonal elements, thereby facilitating the learning of more robust representations. Specifically, for a mini-batch of size $N$, each query feature is associated with a set $\mathcal S=\{(f^i_q, f^j_t), y_{i,j}\}_{j=1}^N$, where $y_{i,j} = 1$ denotes a hard-matched pair (the ground truth), $y_{i,j} = \beta,~\beta \in (0,1)$ represents a soft-matched pair (sharing the same TID), and $y_{i,j} = 0$ indicates an unmatched pair. Then the matching probability of pairs in $\mathcal S$ is computed according to the following softmax function:
\begin{equation} p_{i,j}=\frac{\exp(sim(f^i_{q},f_t^j)/\tau)}{\sum_{k=1}^N\exp(sim(f^i_{q},f_t^k)/\tau)},
\label{probability}
\end{equation}
where $sim(f^i_q,f^j_t)$ is the cosine similarity, and $\tau$ is a temperature hyperparameter that controls the sharpness of the probability distribution. The label distribution, representing the true matching probability, is computed as $q_{i,j}=y_{i,j}/\sum_{k=1}^Ny_{i,k}$ and the global contextual alignment loss from query to target is calculated by:
\begin{equation} \mathcal{L}_{q2t}=\frac{1}{N}\sum_{i=1}^NKL(\mathbf{p_i}\|\mathbf{q_i})=\frac{1}{N}\sum_{i=1}^N\sum_{j=1}^Np_{i,j}\log(\frac{p_{i,j}}{q_{i,j}+\epsilon}),
\end{equation}
where $KL(\cdot||\cdot)$ denotes the KL divergence and $\epsilon$ is used to prevent numerical issues. In the same way, the target-to-query loss $\mathcal{L}_{t2q}$ can be obtained by exchanging $f_q$ and $f_t$ in Equation~\ref{probability}, and the full global contextual alignment loss is as follows:
\begin{equation} 
\mathcal{L}_{gca}=\mathcal{L}_{q2t}+\mathcal{L}_{t2q}.
\end{equation}

\textbf{Local Contextual Reasoning:} Complementary to global contextual alignment, we further propose local contextual reasoning to enable the model to learn more informative features within the context of each triplet, enhancing its capability to handle finer-grained scenes. Unlike masked language/image modeling (MLM/MIM)~\cite{bert,mae}, CoAlign adopts a more lightweight decoder and performs masked feature prediction (MFP)~\cite{mpf} at the latent level, which significantly reduces computational cost. To be specific, for a composed query and its hard-matched target image $(f_q, f_t)$, we first randomly mask out elements along the feature dimension with a probability of 30\%. Following BERT~\cite{bert}, the masked elements are replaced with 10\% random value, 10\% unchanged, and 80\% set to zero, yielding the masked pair $(\tilde{f}_q, \tilde{f}_t)$. Subsequently, a rearrange operation is performed to group and concatenate the features to obtain $[f_q, \tilde{f}_t]\in \mathbb R^{2d}$ and $[f_t, \tilde{f}_q]\in \mathbb R^{2d}$. These concatenated representations are processed through a latent decoder $\Phi$ (a two-layer MLP) that outputs the reconstructed features, which are compared with the original ones using mean squared error (MSE). The local contextual reasoning loss is defined as follows:
% 不要写mse写最小化差异
\begin{equation} 
\mathcal L_{lcr}=\mathbb E_{(f_q,f_t)\sim\mathcal B}[||f_q,\Phi{([f_t, \tilde{f}_q])}||_2^2 + ||f_t,\Phi{([f_q, \tilde{f}_t])}||_2^2].
\end{equation}

The overall training objective $\mathcal L$ is a weighted sum of the global and local terms, where $\gamma$ is a hyperparameter.
\begin{equation} 
\mathcal L = \mathcal L_{gca} + \gamma\mathcal L_{lcr}.
\end{equation}
\subsubsection{Inference Workflow} Given a target image gallery, we first pre-extract features for each image, obtaining $\mathcal V=\{f^j_t\}_{j=1}^{N}$. Then, for a input query $f_q$, the cosine similarity between $f_q$ and the features in $\mathcal V$ is computed, and the top-K images with the highest similarity scores are returned as the retrieval output.

\begin{table}[t]
\small
\centering
\caption{\textbf{Statistics of Common CIR Datasets.} We compare our CIRHS dataset with existing manually annotated and automatically constructed datasets.}
\begin{tabular}{llllc}
\toprule
\multirow{2}{*}{\textbf{Dataset}} & \multirow{2}{*}{\textbf{Domain}} & \multirow{2}{*}{\textbf{Triplets}} & \multirow{2}{*}{\textbf{Images}} & \multirow{1}{*}{\textbf{Text}} \\
 & & & & \multirow{1}{*}{\textbf{length}} \\
\midrule
CIRR~\cite{cirr} & Natural & 36,554 & 21,185 & 59.51 \\
FashionIQ~\cite{fashioniq} & Fashion & 30,132 & 7,988 & 27.13 \\
LaSCo~\cite{case} & Natural & 389,305 & 121,479 & 30.70 \\
WebVid-CoVR~\cite{covr} & Natural & 1,644,276 & 130,559 & 23.36 \\
ST18M~\cite{compodiff} & Synthetic & 18,000,000 & - & - \\
\textbf{CIRHS (Ours)} & Synthetic & 534,758 & 534,758 & 53.17 \\
\bottomrule
\end{tabular}
\label{tbl:stat}
\end{table}

\begin{table*}[t]
\small
\centering
\caption{\textbf{Performance comparison with existing zero-shot CIR methods.} The best results are marked in bold, and the second-best results are underlined. $\dagger$ indicates that the dataset is synthetic.}
\begin{tabular}{c c c c c | c c c c | c c}
\toprule
\multirow{3}{*}{\textbf{Method}} & \multirow{3}{*}{\textbf{Ref.}} & \multirow{3}{*}{\textbf{Pretraining Data}} & \multicolumn{8}{c}{\textbf{Zero-shot Composed Image Retrieval}} \\
& & & \multicolumn{2}{c}{FashionIQ} & \multicolumn{4}{c}{CIRR} & \multicolumn{2}{c}{CIRCO} \\
\cmidrule(lr){4-5} \cmidrule(lr){6-9} \cmidrule(lr){10-11} 
& & & Avg@10 & Avg@50 & R@1 & R@5 & R$_s$@1 & Avg. & mAP@5 & mAP@10 \\
\midrule
\multicolumn{1}{l}{PALAVRA~\cite{personalize}} & \multicolumn{1}{l}{ECCV22} & - & 19.76 & 37.25 & 16.62 & 43.49 & 41.61 & 42.55 & 4.61 & 5.32 \\
\multicolumn{1}{l}{Pic2Word~\cite{pic2word}} & \multicolumn{1}{l}{CVPR23} & CC3M & 24.70 & 43.70 & 23.90 & 51.70 & - & - & - & - \\
\multicolumn{1}{l}{SEARLE~\cite{serealcirco}} & \multicolumn{1}{l}{ICCV23} & ImageNet1k & 27.61 & 47.90 & 24.87 & 52.31 & 53.80 & 53.06 & 11.68 & 12.73 \\
\multicolumn{1}{l}{ContextI2W~\cite{contexti2w}} & \multicolumn{1}{l}{AAAI24} & CC3M & 27.80 & 48.90 & 25.60 & 55.10 &-&-&-&-\\
\multicolumn{1}{l}{KEDs~\cite{ked}} & \multicolumn{1}{l}{CVPR24} & CC3M & 26.80 & 47.90 & 26.40 & 54.80 &-&-&-&-\\
\multicolumn{1}{l}{Slerp+TAT~\cite{slerp}} & \multicolumn{1}{l}{ECCV24} & CC3M & 32.77 & {53.32} & 33.98 & 61.74 &68.55&54.76&18.46& 19.41 \\
\multicolumn{1}{l}{Image2Sentence~\cite{image2sentence}} & \multicolumn{1}{l}{ICLR24} & CC3M & 29.79 & 49.19
 & 30.84 &61.06 &-&-&11.33 & 12.25 \\
\multicolumn{1}{l}{CIReVL~\cite{cirevl}} & \multicolumn{1}{l}{ICLR24} & - & 32.19 & 52.36 & 34.65 & 64.29 & 67.95 & 66.12 & \textbf{26.77} & \textbf{27.59} \\
% \multicolumn{1}{l}{ImageScope~\cite{}} & \multicolumn{1}{l}{WWW25} & - &  \\
\midrule
\multicolumn{11}{c}{\textit{Comparison with methods based on CIR triplet construction}} \\
\midrule
\multicolumn{1}{l}{CoVR-BLIP~\cite{covr}} & \multicolumn{1}{l}{AAAI24} & WebVid-CoVR & 27.70 & 44.63 & \underline{38.48} & {66.70} & {69.28} & {67.99} & 21.43 & 22.33 \\
\multicolumn{1}{l}{CASE~\cite{case}} & \multicolumn{1}{l}{AAAI24} & LaSCo+CoCo & - & - & 35.40 & 65.78 & 64.29 & 65.04 & - & - \\
\multicolumn{1}{l}{CompoDiff~\cite{compodiff}} & \multicolumn{1}{l}{TMLR24} & ST18M$^\dagger$+LAION2B & \underline{39.02} & 51.71 & 26.71 & 55.14 & 64.54 & 59.84 & 15.33 & 17.71 \\
\midrule
\multicolumn{1}{l}{CLIP4CIR~\cite{clip4cir}} & - & CIRHS$^\dagger$ (Ours) & 26.94 & 47.73 & 29.64 & 62.16 & 57.78 & 59.97 & 20.17 & 21.98 \\
\multicolumn{1}{l}{BLIP4CIR~\cite{blip4cir}} & - & CIRHS$^\dagger$ (Ours) & 30.89 & 52.74 & 25.76 & 55.12 & 55.08 & 55.10 & 18.73 & 20.02 \\
\multicolumn{1}{l}{SPRC~\cite{sprc}} & - & CIRHS$^\dagger$ (Ours) & 37.44 & \underline{57.91} & 38.32 & \underline{68.93} & \underline{69.34} & \underline{69.14} & 21.76 & 23.12 \\
\multicolumn{1}{l}{\textbf{CoAlign (Ours)}} & - & CIRHS$^\dagger$ (Ours) & \textbf{39.11} & \textbf{60.29} & \textbf{41.17} & \textbf{71.68} & \textbf{70.65} & \textbf{71.17} & \underline{23.47} & \underline{25.29} \\
\bottomrule
\end{tabular}
\label{zs-sota}
\end{table*}

\begin{table}[t]
\centering
\small
\caption{\textbf{Ablation experiments on each component of CoAlign.} Best results are in bold. To validate the effectiveness of GCA, we additionally introduce the image–text contrastive loss (ITC)~\cite{moco,infonce} for comparison, denoted by $*$.}
\begin{tabular}{c c c c c c c c}
\toprule
\multicolumn{3}{c}{\textbf{Components}} & \multicolumn{3}{c}{CIRR} & \multicolumn{2}{c}{FashionIQ} \\ 
\cmidrule(lr){1-3} \cmidrule(lr){4-6} \cmidrule(lr){7-8}
ITC$^*$ & GCA & LCR & R@5 & R$_{s}$@1 & Avg. & Avg@10 & Avg@50 \\
\midrule
\multicolumn{8}{c}{\textit{Under the zero-shot CIR setting.}} \\
\midrule
\cmark & & & 69.66 & 69.57 & 69.62 & 38.07 & 59.30 \\
 & \cmark & & 71.64 & 69.93 & 70.79 & 39.04 & 59.91 \\
 & \cmark  & \cmark & \textbf{71.68} & \textbf{70.65} & \textbf{71.17} & \textbf{39.11} & \textbf{60.29} \\
\midrule
\multicolumn{8}{c}{\textit{Adopt the supervised CIR setting.}} \\
\midrule
\cmark & - &  & 82.92 & 79.86 & 81.39 & 54.53 & 74.86 \\
\cmark & - & \cmark & \textbf{83.81} & \textbf{80.87} & \textbf{82.34} & \textbf{54.92} & \textbf{75.55} \\
\bottomrule
\end{tabular}
\label{tbl:ablation}
\end{table}

\begin{figure}
\centering
\includegraphics[width=1\columnwidth]{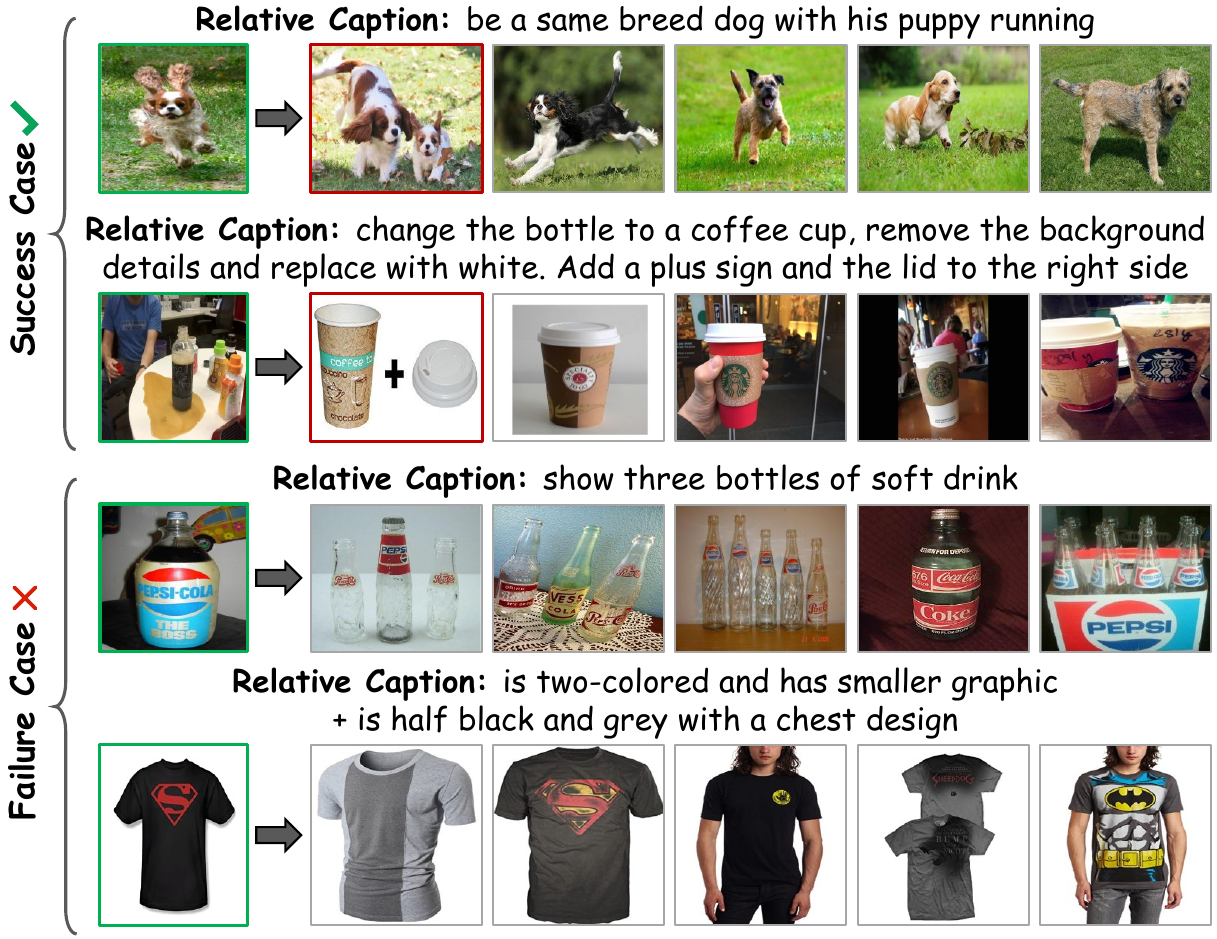}
\caption{\textbf{Qualitative results on CIRR and FahsionIQ.} The reference image and target images are highlighted with green and red outline, respectively.}
\label{fig:case}
\end{figure}

\section{Experiments}
We conduct extensive experiments on three widely-used CIR datasets to demonstrate the effectiveness of our proposed methods.

\subsection{The CIRHS Dataset}
By applying the proposed pipeline, we construct CIRHS, a fully synthetic dataset of 534K CIR triplets. Table~\ref{tbl:stat} summarizes its statistics. Compared with existing manually annotated datasets~\cite{fashioniq, cirr}, CIRHS is significantly larger in scale. Although smaller than WebVid-CoVR~\cite{covr} and ST18M~\cite{compodiff}, CIRHS offers advantages in data quality and diversity. WebVid-CoVR is constrained by the lack of diversity in relative captions (mainly object or scene change). ST18M, on the other hand, based on image editing to generate CIR triplets, suffers from poor generation quality due to unrealistic outputs and visual artifacts. In contrast, CIRHS is explicitly designed to ensure both semantic diversity and high visual quality. Our experiments also show that 534K triplets are sufficient to train strong CIR models. Using CIRHS as the only training corpus, we demonstrate that training CIR models purely on synthetic data is both feasible and effective across three widely-used benchmarks.

\subsection{Experimental Setup}
This section covers the evaluation datasets, metrics, and implementation details of our experiments.
\subsubsection{Evaluation Benchmarks} FashionIQ~\cite{fashioniq} is designed to simulate a realistic online shopping chat interface, focusing on the fashion domain. It is divided into three categories: Dress, Shirt, and Toptee, comprising 30,134 triplets constructed from 77,684 images. CIRR~\cite{cirr} is the first open-domain CIR dataset, containing 21,552 real-life images with human-annotated modified text. CIRCO~\cite{serealcirco} leverages the COCO 2017 unlabeled split~\cite{coco} to construct a benchmark containing 123,403 images, where each query corresponds to multiple ground-truth images, providing a robust evaluation.

\subsubsection{Evaluation Metrics} 
In our experiments on CIRR~\cite{cirr} and FashionIQ~\cite{fashioniq}, Recall@K serves as the primary metric, measuring the likelihood of finding the target image within the top-K retrieved candidates. For CIRR, we additionally report Recall$_s$@K on visually similar subsets, with overall performance summarized as $Avg.=\frac{Recall@5+Recall_s@1}{2}$. On CIRCO~\cite{serealcirco}, where each query has multiple target images, mean Average Precision (mAP@K) is utilized as the principal performance measure.

\subsubsection{Implementation Details} 
(1) During the construction of CIRHS, we utilize eight NVIDIA H800 GPUs. The selected LLM is Qwen2.5-32B~\cite{qwen2}, the T2I-GM is Flux.1-dev~\cite{flux}, and the MLLM for data filtering is Qwen2.5-VL-32B~\cite{qwen2-vl}. For each generated sample, the LLM randomly selects one instruction from six predefined sets to generate a diverse textual quadruple. The T2I-GM then synthesizes a $528\times1056$ resolution image with side-by-side layout, which can be safely cropped into two $512\times512$ images, serving as the reference and target images. Finally, these images and their relative captions are evaluated by the MLLM with scores ranging from $1$ to $10$. A weighted sum is computed across three dimensions using weights $(0.3, 0.2, 0.5)$, and a threshold of $\alpha=7.5$ is applied to filter out low-quality triplets, which accounts for approximately $15\%$. More details is provided in the appendix. (2) For CoAlign, we use BLIP-2~\cite{blip2} with a frozen ViT-G/14~\cite{vit} at an input resolution of 224 pixels. The model is trained on CIRHS for 10 epochs using a batch size of 128 on one NVIDIA H800 GPU. We adopt the AdamW optimizer~\cite{adamw} with an initial learning rate of 5e-6. The hyperparameter $\beta$ and $\gamma$ are set to $0.6$ and $0.4$. For training from scratch on CIRR and FashionIQ, we train for 50 and 30 epochs, with initial learning rates of 1e-5 and 2e-5, using the same batch size of 128.

\subsection{Quantitative Results}
\textbf{Comparison with supervised CIR approaches.} Table~\ref{supervised-sota} presents a comparison of existing supervised CIR methods. Our method, CoAlign, achieves the best overall performance on both FashionIQ and CIRR. Specifically, SPRC focuses on sentence-level prompt optimization but lacks local understanding. CaLa~\cite{cala} aims to capture fine-grained query-target relations but suffers from suboptimal global alignment. In contrast, CoAlign adopts hybrid contextual alignment, jointly optimizing both global and local objectives in a simple yet effective manner.

\textbf{Comparison with ZS-CIR methods.} Table~\ref{zs-sota} compares existing zero-shot CIR methods. Our approach is the only one trained solely on synthetic triplets while achieving strong performance. Among methods based on CIR triplet construction, it outperforms all others across all metrics. Notably, CIRHS is compatible with any CIR framework, e.g., SPRC and CLIP4CIR also perform well when trained on it. On CIRCO, our method ranks second. This is primarily due to the large visual discrepancies between reference and target images inherent in CIRCO, where retrieval relies heavily on the relative caption. This reliance deviates from the original intent of CIR and makes it more favorable to training-free methods such as CIReVL. However, the complex architectures of such methods hinder domain-specific fine-tuning, whereas our approach supports it, offering greater flexibility.

\begin{figure}
\includegraphics[width=0.96\columnwidth]{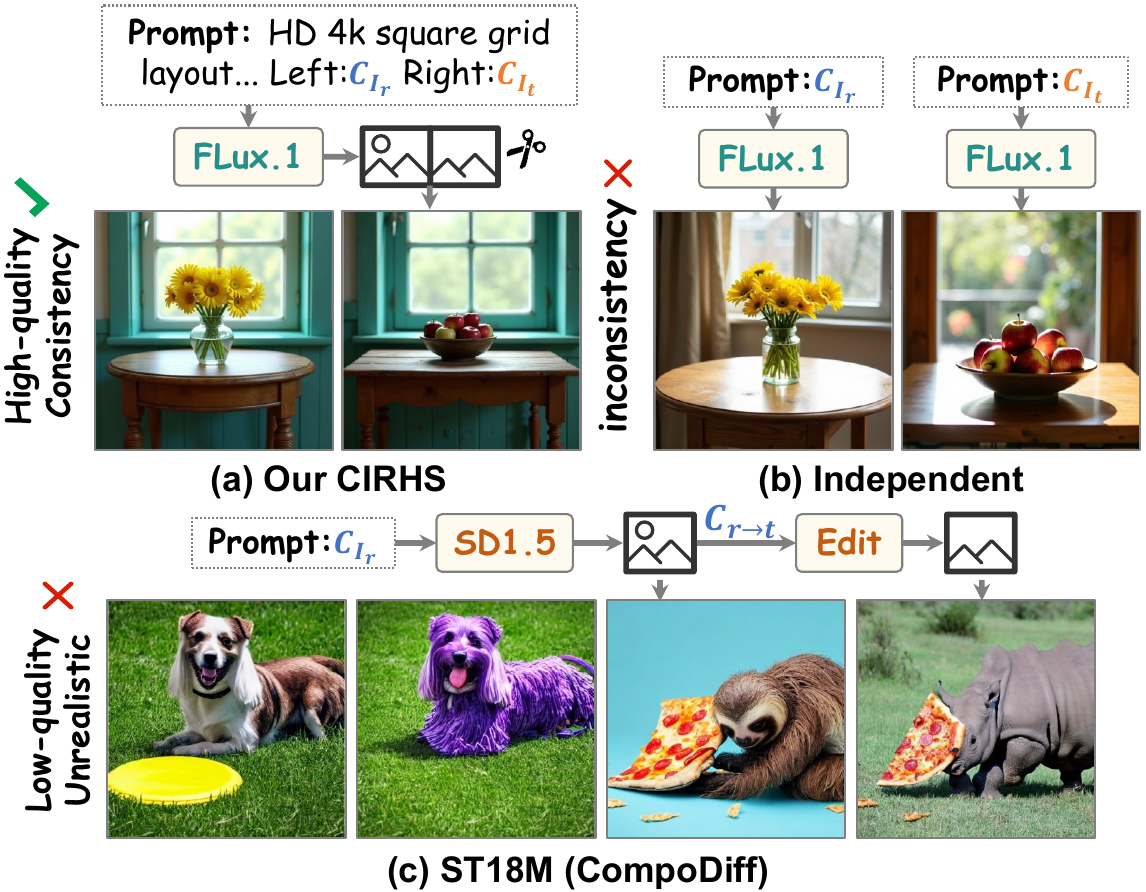}
\caption{\textbf{Comparison of three paradigms for synthesizing image pairs.} Compared to the other two approaches (b) and (c), our approach (a) is superior in both generation quality and consistency.}
\label{fig:example}
\end{figure}

\subsection{Qualitative Results}
Figure~\ref{fig:case} showcases qualitative predictions. (1) The top two rows demonstrate strong multimodal reasoning capacities of our method, including object composition and fine-grained semantic comprehension. (2) The bottom two rows analyze failure cases. Both CIRR and FashionIQ contain numerous false negatives, for instance, the first two predictions in the third row are actually correct. The fourth row highlights an ambiguous relative caption, a common issue in the FashionIQ dataset. Additionally, our model tends to favor outputs with similar backgrounds, which can be attributed to the consistency constraints during dataset construction.

\subsection{Ablation Study}
In this section, we conduct ablation studies to validate the contribution of each component in our proposed method.

\textbf{Our CoAlign Model}. We train multiple versions of our model on CIRR and FashionIQ, as shown in Table~\ref{tbl:ablation}. Under the zero-shot CIR setting, we additionally introduce the widely used image–text contrastive loss (ITC)~\cite{moco,infonce}, which optimizes only the diagonal of the similarity matrix. This allows us to evaluate the effectiveness of Global Contextual Alignment (GCA) in learning from broader cross-modal contexts. Experimental results demonstrate the effectiveness of GCA, and show that combining it with Local Contextual Reasoning (LCR) for fine-grained alignment can further improve retrieval performance. Results under supervised training also support this finding. Note that in the supervised setting on CIRR and FashionIQ, each composed query corresponds to only one target image, making ITC and GCA functionally equivalent.

\textbf{Results with Different Datasets.} To evaluate the effectiveness of our proposed triplet synthesis pipeline, we train the CoAlign model on multiple datasets under the same configuration and report the performance on CIRR and FashionIQ. Note that we randomly sample 100K triplets from each dataset for training to reduce computational cost. As shown in Table~\ref{tbl:dataset}, the model trained on CIRHS outperforms that trained on the real-world dataset WebVid-CoVR~\cite{covr}. We attribute this to the semantic diversity introduced by the LLM during triplet construction. Compared to ST18M~\cite{compodiff}, CIRHS significantly improves the quality of the synthesized triplets, particularly in producing more photorealistic and high-quality reference and target images, which leads to better performance and stronger generalization to real-world scenarios. We also evaluate an alternative generation method where captions $C_{I_r}$ and $C_{I_t}$ are used independently as prompts for the T2I-GM (denoted as \textit{Independent} in Table~\ref{tbl:dataset}). Due to the lack of consistency, this setting is suboptimal for building CIR triplets. In addition, Figure~\ref{fig:example} compares three image-pair synthesis paradigms, providing a visual explanation of the above conclusions. Finally, we verify the effectiveness of the third-stage on data filtering. As shown in Table~\ref{tbl:dataset}, filtering out low-quality samples further improves performance, for example, \textbf{+0.53\%} Recall@5 and \textbf{+1.57\%} R$_s$@1 on CIRR, highlighting the importance of our data filtering strategy.

\begin{figure}[t]
\includegraphics[width=1\columnwidth]{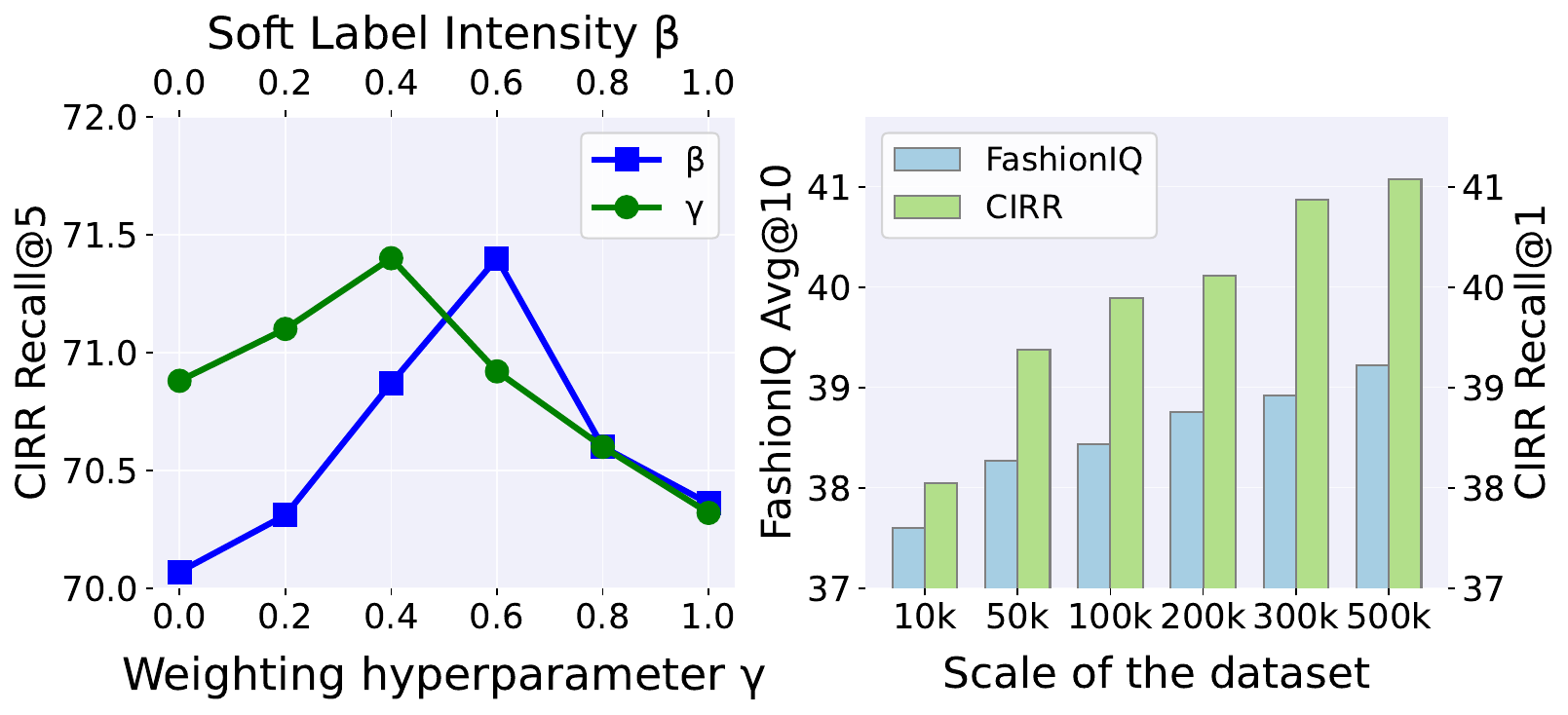}
\caption{\textbf{Hyperparameter and data scale analysis.} Left: Sensitivity analysis of CoAlign on different hyperparameters. Right: Impact of data scale on zero-shot performance.}
\label{fig:hyper}
\end{figure}

\textbf{Hyperparameter and Data Scale Analysis.} (1) The left side of Figure~\ref{fig:hyper} illustrates the effect of two key hyperparameters in our proposed CIR framework CoAlign, namely $\beta$ and $\gamma$, evaluated on CIRR under the zero-shot setting with Recall@5. For the soft label intensity $\beta$, we vary its value from 0 to 1 with a step size of 0.2. The performance first improves and then declines as $\beta$ increases, with the optimal value observed at $\beta=0.6$. For the weighting parameter $\gamma$, which balances GCA and LCR, we apply the same sampling strategy. A similar trend is observed, with a different performance peaking at $\gamma=0.4$. (2) The right side of Figure~\ref{fig:hyper} presents the impact of training data scale, reporting zero-shot results on CIRR and FashionIQ. We observe that with a small data size (e.g., 10K), the model performs poorly, indicating insufficient knowledge acquisition. As the dataset size increases, performance improves and saturates around 300K samples, suggesting that our CIRHS dataset with 534K triplets is enough for training strong CIR models.

\begin{table}[t]
\centering
\small
\caption{\textbf{Results with Different Datasets.} We sample 100K triplets from each dataset for training, and evaluate the models on zero-shot metrics over CIRR and FashionIQ.}
\begin{tabular}{l c c c c c c c}
\toprule
\multirow{2}{*}{\textbf{Dataset}} & \multirow{2}{*}{\textbf{Filtering}} & \multicolumn{2}{c}{CIRR} & \multicolumn{2}{c}{FashionIQ} \\ 
\cmidrule(lr){3-4} \cmidrule(lr){5-6}
 &  & R@5 & R$_{s}$@1 & Avg@10 & Avg@50 \\
\midrule
WebVid-CoVR~\cite{covr} & - & 67.28 & 70.19 & 36.45 & 57.66 \\
LaSCo~\cite{case} & - & 63.24 & 65.77 & 33.12 & 54.87 \\
ST18M~\cite{compodiff} & - & 60.00 & 57.59 & 30.60 & 51.00 \\
Independent & \cmark & 70.17 & 68.97 & 37.19 & 58.47 \\
\midrule
\multirow{2}{*}{\textbf{CIRHS (Ours)}} & \xmark & 70.65 & 68.75 & 37.24 & \textbf{59.47} \\
 & \cmark & \textbf{71.18} & \textbf{70.32} & \textbf{37.76} & 59.28 \\
\bottomrule
\end{tabular}
\label{tbl:dataset}
\end{table}

\section{Discussion}
While our method significantly improves the quality of automatically synthesized CIR triplets and is the first to demonstrate the feasibility of training robust CIR models purely on synthetic data, it still has certain limitations: (1) We use Flux.1-dev as the T2I-GM, which is trained with aesthetic-aware objectives that bias it toward generating overly polished or aesthetically pleasing images. This causes a distribution gap between the synthesized images and casually captured photos found in benchmarks (e.g. CIRR and CIRCO), potentially compromising model performance. Reducing this gap is a direction for our future work. (2) Existing methods, including ours, do not consider retrieval across varying resolutions or image clarity, although this could be important in practical scenarios. A potential solution is to fine-tune the T2I-GM at multiple resolutions to generate images for triplet construction. (3) Current CIR benchmarks also present limitations. For instance, retrieval in CIRCO relies heavily on the relative caption, and the reference image may contribute little, or it may even negatively affect performance. This contradicts the original intent of CIR. Future efforts are needed to refine and redesign CIR benchmarks for better alignment .

\section{Conclusion}
In this paper, we propose a scalable pipeline for the automatic synthesis of high-quality triplets, addressing key limitations of previous triplet construction approaches such as limited image fidelity and semantic diversity. Using this pipeline, we construct a large-scale synthetic dataset, referred to as Composed Image Retrieval on High-quality Synthetic Triplets (CIRHS). The pipeline leverages an LLM to generate diverse and semantically rich prompts, which guide a T2I-GM to produce consistent image pairs. These image pairs are subsequently filtered and reorganized to form the triplet dataset. In addition, we introduce a new CIR framework, Hybrid Contextual Alignment (CoAlign), which jointly optimizes from both global and local perspectives within a broader context. By utilizing the CIRHS dataset, CoAlign achieves outstanding zero-shot performance on three standard benchmarks. To the best of our knowledge, this is the first approach to train CIR models entirely on synthetic data while achieving strong performance. Under the supervised setting, CoAlign further outperforms all state-of-the-art CIR methods, confirming the effectiveness of our retrieval framework.

\bibliographystyle{ACM-Reference-Format}
\bibliography{sample-base}
\end{document}